\pdfoutput=1

\documentclass[11pt]{article}

\usepackage[]{acl}

\usepackage{times}
\usepackage{latexsym}

\usepackage[T1]{fontenc}

\usepackage[utf8]{inputenc}

\usepackage{microtype}

\usepackage{inconsolata}

\usepackage{booktabs}
\usepackage{graphicx}
\usepackage{xspace}
\usepackage{enumitem}
\newcommand{\method}{\texttt{DiagGPT}\xspace}

\title{DiagGPT: An LLM-based and Multi-agent Dialogue System with Automatic Topic Management for Flexible Task-Oriented Dialogue}

\author{Lang Cao \\
  University of Illinois Urbana-Champaign \\
  Department of Computer Science \\
  \texttt{langcao2@illinois.edu} \\}

\begin{document}
\maketitle
\begin{abstract}
A significant application of Large Language Models (LLMs), like ChatGPT, is their deployment as chat agents, which respond to human inquiries across a variety of domains. While current LLMs proficiently answer general questions, they often fall short in complex diagnostic scenarios such as legal, medical, or other specialized consultations. These scenarios typically require Task-Oriented Dialogue (TOD), where an AI chat agent must proactively pose questions and guide users toward specific goals or task completion. Previous fine-tuning models have underperformed in TOD and the full potential of conversational capability in current LLMs has not yet been fully explored. In this paper, we introduce DiagGPT (Dialogue in Diagnosis GPT), an innovative approach that extends LLMs to more TOD scenarios. In addition to guiding users to complete tasks, DiagGPT can effectively manage the status of all topics throughout the dialogue development. This feature enhances user experience and offers a more flexible interaction in TOD. Our experiments demonstrate that DiagGPT exhibits outstanding performance in conducting TOD with users, showing its potential for practical applications in various fields.
\end{abstract}

\section{Introduction}
\begin{figure}[t!]
\centerline{
\resizebox{.5\textwidth}{!}{
\includegraphics{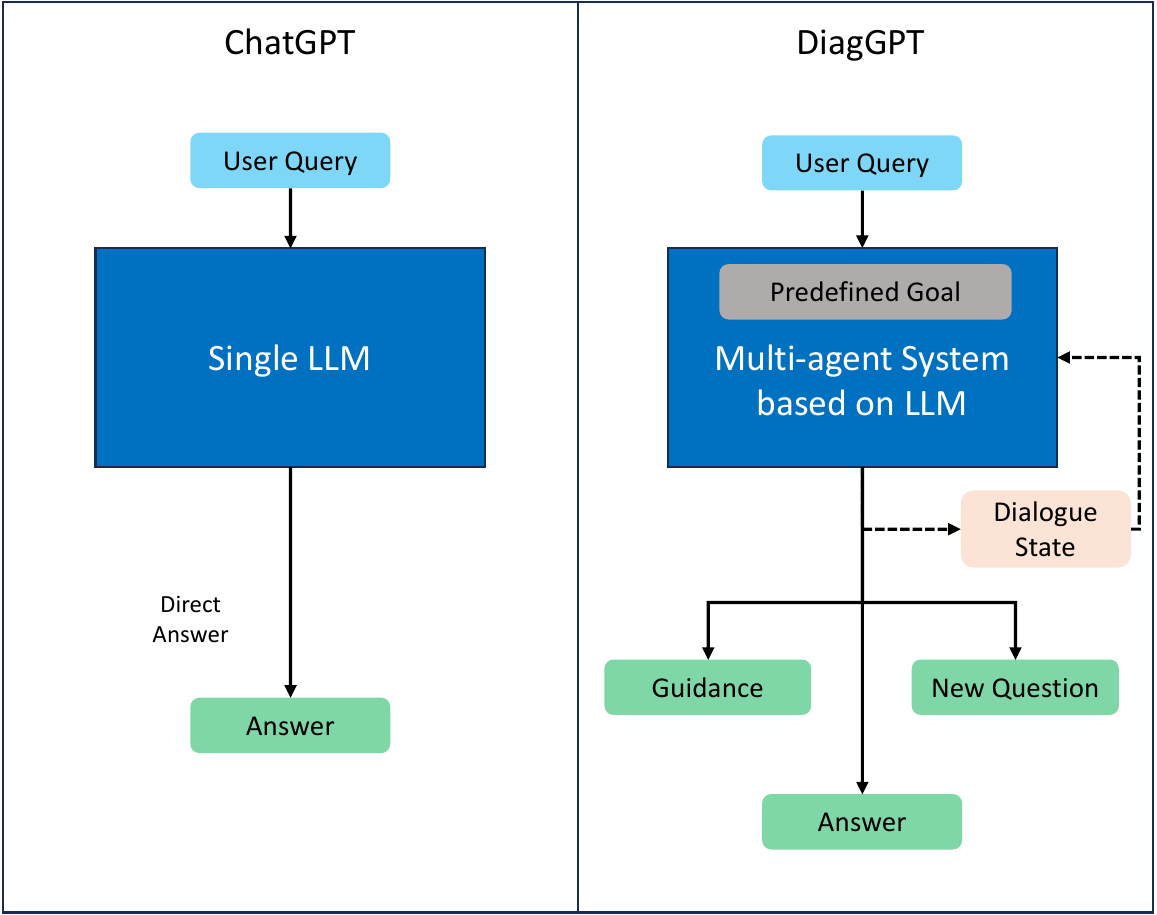}
}}
\caption{The main difference between ChatGPT and \method. While ChatGPT directly answers user questions, \method not only provides answers of the same quality but also has the ability to proactively ask questions, guide users, and maintain dialogue state internally.}
\label{fig:overview}
\end{figure}


Large language models (LLMs), such as ChatGPT, have demonstrated remarkable performance on various natural language processing (NLP) tasks \cite{NEURIPS2020_1457c0d6, chowdhery2022palm, wei2022emergent, openai2023gpt4}. In many cases, OpenAI GPT-4 even outperforms human performance \cite{openai2023gpt4}. With the use of prompt engineering techniques (e.g., chain-of-thought prompting \cite{NEURIPS2020_1457c0d6, wei2022chain}, in-context learning \cite{NEURIPS2020_1457c0d6, xie2022an, min-etal-2022-rethinking}, etc.), we can improve reasoning and understanding of LLMs to complete complex tasks in our daily life. LLMs have attracted enormous attention from both academia and industry, inspiring more people to build useful applications based on them.

One popular application of LLMs is in chatbots, which build conversational systems around these models. ChatGPT\footnote{https://openai.com/blog/chatgpt} is a successful example of such an application, where LLMs have the ability to analyze context and respond to user queries based on knowledge derived from extensive training data. By supplementing its background knowledge and providing context and appropriate prompts, ChatGPT has been able to form robust question-answering models for specialized fields. It can understand users' questions and provide precise answers effectively.

However, dialogue scenarios in our daily life can be more complex. For instance, in specialized professional consultation scenarios like legal or medical diagnosis, the chat agent needs to consider the user's unique situation or information. In the process of obtaining user information, the interactive experience provided by the agent is also crucial. The system needs to proactively ask questions. Therefore, we need a consultation process from the chat agent that better simulates real medical experts and legal professionals. The chat agent should conduct question-answering, topic management, and guiding users towards specific goals or task completion. This type of dialogue is known as Task-Oriented Dialogue (TOD). There are usually some predefined goals in a conversation. TOD helps users achieve their specific goals, focusing on understanding users, tracking states, and generating next actions \cite{balaraman-etal-2021-recent}. It is substantially different from light-conversational or open-domain dialogue scenarios. Despite much research in this area, it remains challenging due to issues such as a lack of training data, inefficiency, and drawbacks of fine-tuning small models, including an inability to fully understand user meaning and poor generative performance. Models from the existing research on this topic is not robust and universal. For example, fine-tuning models require a lot of data to train and are difficult to transfer to other scenarios. On the other side, although LLMs have a wide range of knowledge and the quality of their answer is far beyond that of fine-tuning models, traditional LLMs no longer meet needs of TOD and cannot effectively manage complex dialogue logic. Because they maintain a simple memory base and can only handle linear interaction.

Recent advancements have focused on using LLMs as agents to construct multi-agent systems or to teach LLMs how to use tools to accomplish more complex tasks \cite{schick2023toolformer, shen2023hugginggpt}. These systems typically have a core agent that controls the entire task process. A prominent example is AutoGPT\footnote{https://github.com/Significant-Gravitas/Auto-GPT}, which employs multiple GPT models to strategize the responsibilities of each agent in order to split complex tasks and then complete them. In such multi-agent systems, the key lies in the division of tasks and the interaction between agents. By organizing multiple LLMs and instruct them to collaborate, we have the opportunity to tackle many complex tasks which one LLM cannot do well.

Fine-tuning models fall short in terms of scenario transfer ability and massive training data requirements, while a single large language model is not proficient in dialogue state tracking and management. However, we can leverage the strong knowledge background of LLMs and employ a multi-agent framework to incorporate the dialogue state tracking and management ideas from fine-tuning models. Thus, we can construct a multi-agent dialogue system that fulfills the requirements of flexible task-oriented dialogue.

Motivated by these considerations, we propose \method in this paper. \method stands for \textbf{Dia}lo\textbf{g}ue in \textbf{Diag}nosis model based on \textbf{GPT}-4. This is a multi-agent AI system, which has automatic topic management ability to enhance its utility for task-oriented dialogue in a more flexible way. Instead of mechanically and rigidly guiding the user through the task in traditional task-oriented dialogue, we can enhance the interactive experience in the development of dialogue, and the dialogue is more smooth. This is more in line with actual situations and real conversations. In summary, our AI system \method possesses the following features:
\begin{itemize}[leftmargin=*, itemsep=0pt, labelsep=5pt]
    \item \textbf{Question Answering} It is a basic feature of traditional LLM-based conversational systems. LLMs possess a wide range of knowledge and can provide high-quality answers to various questions. we build upon and retain this essential capability of LLMs.
    \item \textbf{Task Guidance} The system is designed to guide users towards a specific goal and assist them in accomplishing the task throughout the dialogue progression. This is achieved by advancing a sequence of predefined topics throughout the dialogue.
    \item \textbf{Proactive Asking} The system has the ability to proactively pose questions based on a predefined checklist, thereby collecting necessary information from users.
    \item \textbf{Topic Management} The system is capable of automatically managing topics throughout the dialogue, tracking topic progression, and effectively engaging in discussions centered around the current topic. It performs well in managing various topic changes in complex dialogues.
    \item \textbf{Versatile} Our system is directly based on LLMs. It can perform well in various scenarios without requiring any training data, a capability that previous fine-tuning models lack. This system can be easily applied to multiple cases by defining specific predefined goals and supplementing functions to support them.
\end{itemize}
Given these features, \method can meet the aforementioned needs and better engage in professional consultation conversations with users in complex scenarios. Our main contribution is to make traditional LLM-based conversational systems smarter in flexible task-oriented dialogue. We build on the strong knowledge of LLMs and give them more interactive capabilities. Therefore, \method can function like a more intelligent and more professional chatbot. 



\section{Related Work}

\noindent \textbf{Task-Oriented Dialogue} systems assist users in achieving specific goals, focusing on understanding users, tracking states, and generating subsequent actions. Task-Oriented Dialogue (TOD) is a task in which the goal is to accomplish a specific objective. Recent work primarily focusing on fine-tuning small models. \cite{wen-etal-2017-network} introduce a neural network-based text-in, text-out end-to-end trainable task-oriented dialogue system along with a new way of collecting dialogue data based on a novel pipe-lined Wizard-of-Oz framework. \cite{wu-etal-2019-transferable} propose a Transferable Dialogue State Generator (TRADE) that generates dialogue states from utterances using copy mechanism, facilitating transfer when predicting (domain, slot, value) triplets not encountered during training. \cite{feng-etal-2023-schema} propose SG-USM, a novel schema-guided user satisfaction modeling framework. It explicitly models the degree to which the user’s preferences regarding the task attributes are fulfilled by the system for predicting the user’s satisfaction level. \cite{liu-etal-2023-one} propose a framework called MUST to optimize TOD systems via leveraging Multiple User Simulator. \cite{bang-etal-2023-task} propose an End-to-end TOD system with Task-Optimized Adapters which learn independently per task, adding only small number of parameters after fixed layers of pre-trained network. All these methods require a considerable amount of data for training and have not yet attained a performance level that is ideal for real-world applications. \\

\noindent \textbf{Conversational Systems with LLMs} have become popular as the robust capabilities of LLMs have been recognized. \cite{hudecek2023llms} evaluated the conversational ability of LLMs and found that, in explicit belief state tracking, LLMs underperform compared to specialized task-specific models. This suggests that simple LLMs do not have the ability to achieve task-oriented dialogue. \cite{liang2023c5} proposed an interactive conversation visualization system called C5, which includes Global View, Topic View, and Context-associated QA View to better retain contextual information and provide comprehensive responses. From another perspective, \cite{zhang2023ask} proposed the Ask an Expert framework in which the model is trained with access to an expert whom it can consult at each turn. This framework utilizes LLMs to improve fine-tuning small models in TOD. There is minimal work on improving the conversational ability of LLMs. To the best of our knowledge, we are the first to successfully use off-the-shelf LLMs in a multi-agent framework to build a task-oriented dialogue system.

\section{Methodology}


\subsection{\method Framework}
\method is a multi-agent and collaborative system composed of several modules: \textit{Chat Agent}, \textit{Topic Manager}, \textit{Topic Enricher}, and \textit{Context Manager}. Each module is a LLM with specific prompts that guide their function and responsibility. Among these modules, the \textit{Topic Manager} is the most important as it tracks the dialogue state and automatically manages the dialogue topic.

\begin{figure*}[t!]
\centerline{
\resizebox{0.9\textwidth}{!}{
\includegraphics{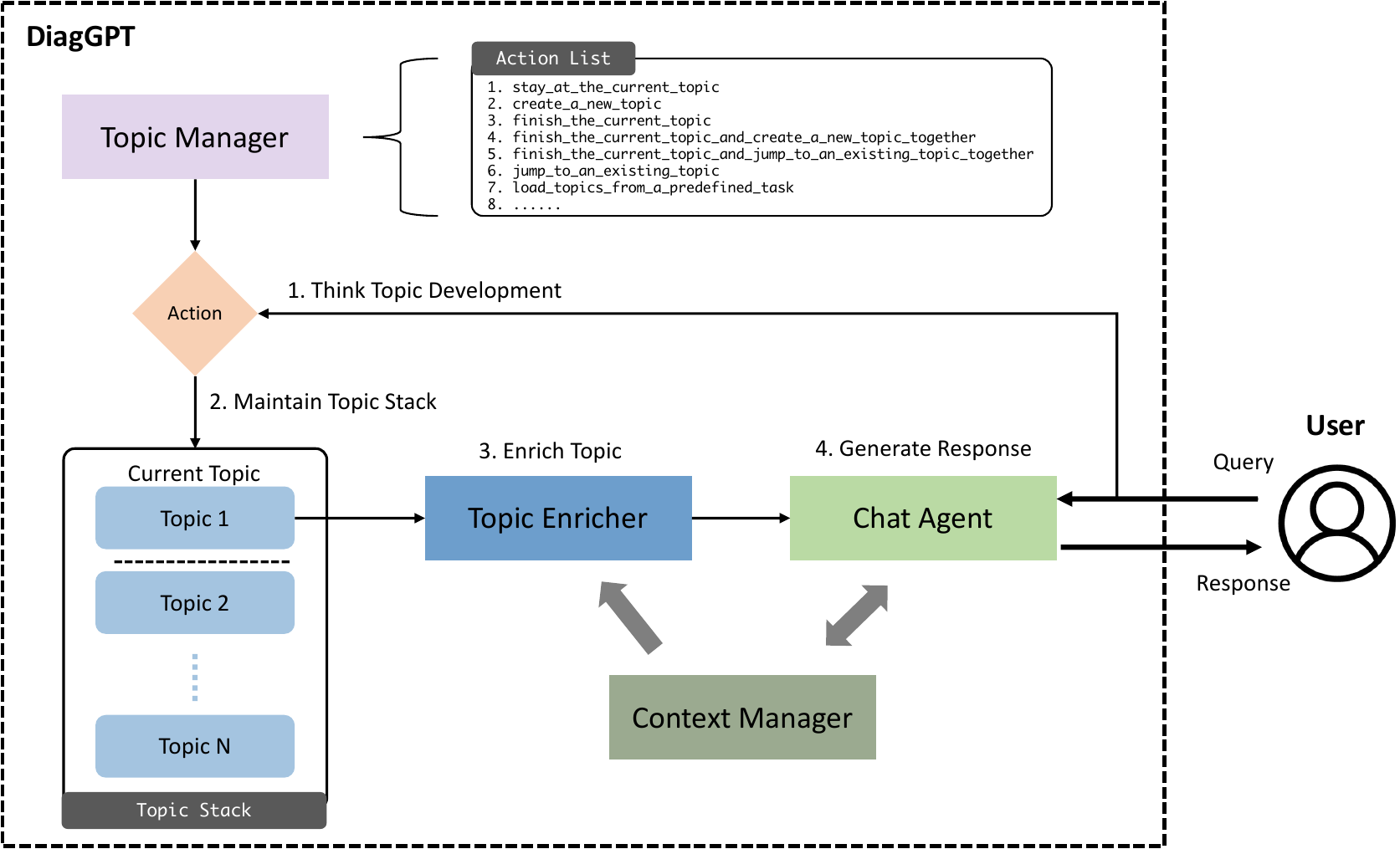}
}}
\caption{The framework of \method. The workflow of \method consists of four stages: Thinking Topic Development, Maintaining Topic Stack, Enriching Topic, Generating Response.}
\label{fig:diaggpt}
\end{figure*}

As shown in Figure~\ref{fig:diaggpt}, the workflow of \method consists of four stages: 1) Thinking Topic Development: \textit{Topic Manager} obtain the user query, then analyze and predict the topic development in current round of dialogue; 2) Maintaining Topic Stack: maintain the topic stack of the entire dialogue according to action commands from \textit{Topic Manager}; 3) Enriching Topic: retrieve the current topic and enrich it based on dialogue context; 4) Generating Response: based on specific guidance prompt and combined it with enriched topic and context to generate response for users.

Besides, we define a topic as the main subject of a round of dialogue, which determines the primary focus of communication. We also define a task as a specific goal that needs to be completed in a task-oriented dialogue. After going through all the predefined topics in a dialogue, this specific task should be accomplished.

\subsection{Thinking Topic Development}
\textit{Topic Manager} serves as the main module in \method and is responsible for determining the topic development based on the user's query. In each round of dialogue, the system needs to adjust the current dialogue topic before providing its response. Therefore, the user's query is first fed into \textit{Topic Manager}.

The input to the \textit{Topic Manager} includes the current user query, action list, the current status of the topic stack, and the chat history. It is logical for an AI agent to analyze and predict the topic development based on this information. Of particular importance is the action list stored in the \textit{Topic Manager}. This action list contains various actions that serve as tools for the \textit{Topic Manager} to execute. The \textit{Topic Manager} has knowledge about the details of each action, how to plan and execute them. Each action corresponds to a program function that executes a specific command. In Python, we use decorator functions to implement this. Whenever \textit{Topic Manager} receives a user query, it analyzes all the available information and decides which action to execute based on the prompts associated with each action.

With the strong understanding and reasoning abilities of LLMs, this AI agent can accurately comprehend the user's intentions and help to effectively engage in communication with the user.

\subsection{Maintaining Topic Stack}
After obtaining the output of the action from the \textit{Topic Manager}, the system will execute the corresponding command to process and control the topic change, which involves maintaining the topic stack.

The topic stack is a data structure in this system that stores and tracks the dialogue state. Stack management is a kind of simulation of dialog state management in both short-term and long-term context. In our system, we concretize the dialogue state of the LLM and explicitly store it, allowing the system to perform better over the long term context and avoid ambiguity and forgetfulness. We consider the progress of a dialogue to have multiple stages or states, and these states follow a first-in, first-out (FIFO) order, which can be effectively modeled using a stack. Although we refer to this topic storage structure as a stack, this component does not strictly adhere to the FIFO rule. This is due to the presence of complex dialogue logic, and FIFO is merely the most common form of topic development. We will provide a detailed description of some operations of this structure in the following paragraphs.

In a diagnosis scenario, a consultant typically has a checklist stored in their mind. In many common cases, if users do not propose any new questions, the dialogue development will follow this checklist. After going through all the items in the checklist, the consultant can provide reports and comprehensive analysis to the users and complete the specific task. The action \textit{load topics from a predefined task} is designed to facilitate this process. When the function decorated by this action prompt is executed, a list of topics from the checklist, will be loaded into the topic stack.

\begin{figure}[t!]
\centerline{
\resizebox{.4\textwidth}{!}{
\includegraphics{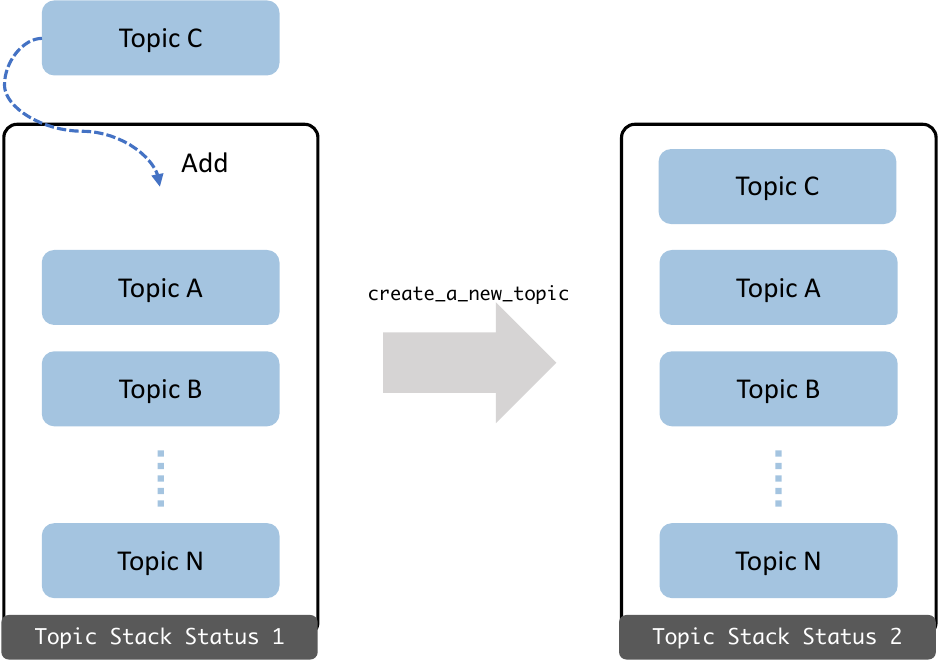}
}
}
\caption{The action of \textit{create a new topic}.}
\label{fig:action_create}
\end{figure}

\begin{figure}[t!]
\centerline{
\resizebox{.4\textwidth}{!}{
\includegraphics{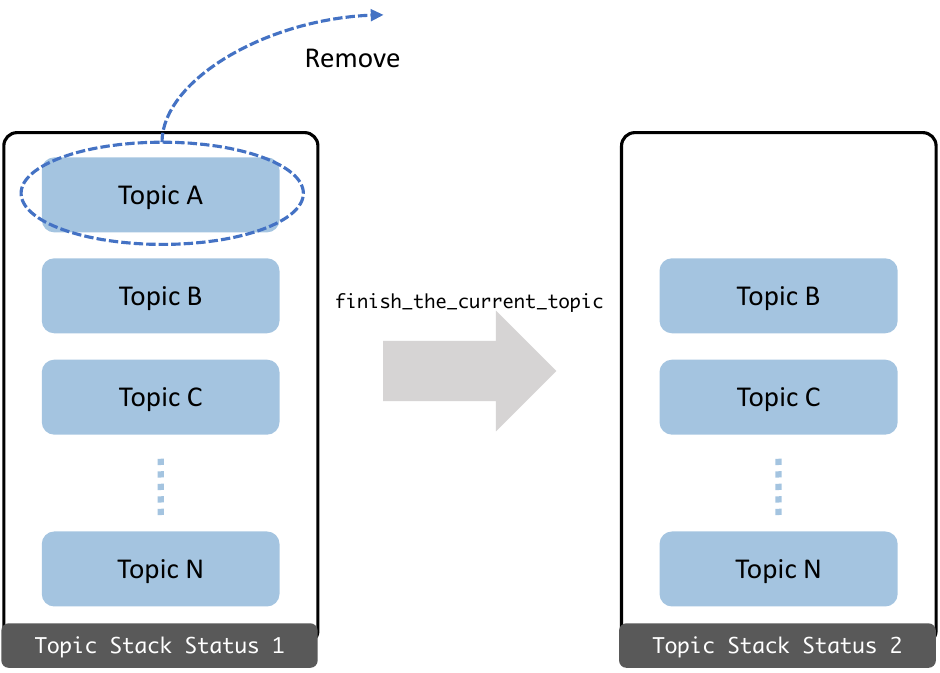}
}
}
\caption{The action of \textit{finish the current topic}.}
\label{fig:action_finish}
\end{figure}

Furthermore, there are other actions commonly used to manipulate the topic stack. These actions include \textit{create a new topic}, \textit{finish the current topic}, and \textit{stay at the current topic}. The \textit{create a new topic} action, as shown in Figure~\ref{fig:action_create}, adds a new topic to the stack when the user wants to start a new topic. The \textit{finish the current topic} action, shown in Figure~\ref{fig:action_finish}, removes the top topic from the stack when the user no longer wishes to discuss it or the system considers this topic to be closed. The \textit{stay at the current topic} action indicates that the system determines that it still requires information and needs continuous discussing the current topic, so the topic stack does not change at all. These three basic operations cover most topic change scenarios. Since we only allow one-step reasoning for LLMs, they must select and execute only one action. Other actions are complex changes based on these three basic operations.

\begin{figure}[t!]
\centerline{
\resizebox{.4\textwidth}{!}{
\includegraphics{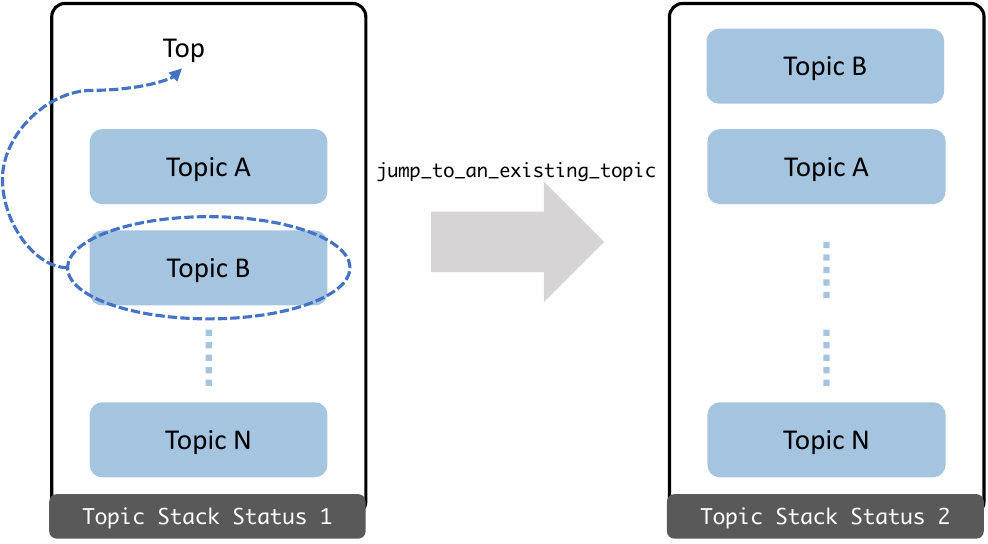}
}
}
\caption{The action of \textit{jump to an existing topic}.}
\label{fig:action_jump}
\end{figure}

The \textit{jump to an existing topic} action, illustrated in Figure~\ref{fig:action_jump}, allows the user to retrieve and prioritize a previous topic from the stack.

We apply a action list to allow this system to interact with users in a more flexible way. It can avoid mechanically and rigidly guiding the user through the task in traditional task-oriented dialogue. This action list can also be expanded to accommodate more complex scenarios in task-oriented dialogues. We have also implemented a mechanism to automatically remove redundant topics. After several rounds of dialogue, if a newly generated topic is not recalled, it will be removed. However, this removal does not affect any predefined topics from checklist.

\begin{table*}[]
\centering
\begin{tabular}{l|cccc|cccc}
\hline
                        & \multicolumn{4}{c|}{Human Evaluation}                      & \multicolumn{4}{c}{LLM Evaluation}                         \\ \cline{2-9} 
                        & RC $\downarrow$           & CR $\uparrow$           & SR $\uparrow$           & CS $\uparrow$            & CR $\uparrow$           & SR $\uparrow$           & RQ $\uparrow$           & CS $\uparrow$            \\ \hline
ChatGPT (\textit{gpt-3.5-turbo}) & 9.0          & 0.8          & 0.4          & -             & 0.8          & 0.7          & 7.8          & -             \\
ChatGPT (\textit{gpt-4})         & 7.7          & 1.0          & 1.0          & 5.0           & 1.0          & 1.0          & 9.0          & 8.5           \\
DiagGPT (Ours)          & \textbf{7.0} & \textbf{1.0} & \textbf{1.0} & \textbf{15.0} & \textbf{1.0} & \textbf{1.0} & \textbf{9.0} & \textbf{11.5} \\ \hline
\end{tabular}
\caption{Quantitative experimental results. We compare our method \method with ChatGPT (\textit{gpt-turbo-3.5} and \textit{gpt-4}) across 20 diverse dialogue scenarios, assessed by Round Count (RC), Completion Rate (CR), Response Quality (RQ), and Comparison Score (CS).}
\label{tab:exp}
\end{table*}

\subsection{Enriching Topic}
We select the top item in the topic stack as the current topic. However, it cannot be directly used as a chat topic for the \textit{Chat Agent} to interact with the user. Topics in the topic stack are simple and convenient to store, whereas topics for the practical use of the \textit{Chat Agent} need to include more information. Without a topic enricher, it is difficult for the \textit{Chat Agent} to directly understand the main objective in the current round of dialogue.

The \textit{Topic Enricher} is designed to bridge this gap and assist in better organizing the language for use. We initially categorize the topic into \textit{Ask user} and \textit{Answer user}. Typically, newly generated topics fall under \textit{Answer user}, while predefined topics are categorized as \textit{Ask user}. This design helps the system determine whether to answer a user's question or ask a question in the current round of dialogue. The \textit{Topic Enricher} takes the output of the \textit{Context Manager} and the current topic to enrich it into a topic that contains more context information. This enriched topic is then provided to the \textit{Chat Agent}.

\subsection{Generating Response}
With the final topic, the \textit{Chat Agent} recognizes it as the primary topic in this round of dialogue. Thus, with context from the \textit{Context Manager}, it can finally generate responses for users. In addition, as shown in Figure~\ref{fig:prompt_main}, some retrieved background knowledge, instructions, and encouragements will also be added into prompts here to further improve the response quality.

\section{Experiments}
\subsection{Setups}
We conduct both quantitative experiments and qualitative experiments to demonstrate the performance of \method. Details of the system implementation can be found in Appendix~\ref{sec:imple}.



The quantitative evaluation of \method is challenging due to the difficulty in quantifying the system's ability to manage various topics that develop in a dialogue. Therefore, we employ quantitative experiments to only assess the system's basic capabilities in topic advancement and guidance. We do not evaluate our system using task-oriented dialogue datasets such as DialoGLUE \cite{EvalAI, moghe-etal-2023-multi3nlu}. As \method is an open system, and it does not require training unlike fine-tuning models. The output of our system is conditioned solely on prompts. Moreover, these traditional TOD datasets focuses on specific subtasks (intent prediction, slot filling, etc.) that were considered necessary for TOD, while LLM-based systems circumvents these specific tasks in a more end-to-end fashion. Therefore, evaluating our system using traditional methods is challenging.

To quantitatively evaluate our system, we develop a new dataset called LLM-TOD, which encompasses 20 different topics in task-oriented dialogue scenarios (details are shown in Appendix~\ref{sec:dataset}). Each data of a TOD scenario includes a checklist of items to go through, finally achieving a specific goal. This checklist and the goal are input into the LLMs to complete a TOD. Besides, we develop a new evaluation schema based on LLMs for TOD tasks. An LLM-based user simulator is also employed to interact with the dialogue system. The results are evaluated using a combination of human assessment and another LLM. During the LLM evaluation, the inputs include basic TOD information and the complete chat history.

On the other hand, qualitative experiments more effectively demonstrate the capabilities and performance of \method. These experiments show that \method can conduct task-oriented dialogues more flexibly.



\subsection{Quantitative Experiment Results}
We evaluate the performance through both human evaluation and LLM evaluation. We compared our method with two ChatGPT baselines: \textit{gpt-3.5-turbo} and \textit{gpt-4}. The evaluation involved the following metrics:

\begin{itemize}[leftmargin=*, itemsep=0pt, labelsep=5pt]
    \item \textbf{Round Count (RC)}: The average number of rounds required to complete a task-oriented dialogue. This metric can be seen as a measure of system efficiency, where lower values are preferable.
    \item \textbf{Completion Rate (CR)}: The percentage of completed checklist items out of the total number of items. An ideal system should complete all items on the checklist.
    \item \textbf{Success Rate (SR)}: The indicator of whether the system successfully completed the task to achieve the intended goal. This metric is  a value of 0 or 1 for each task-oriented dialogue.
    \item \textbf{Response Quality (RQ)}: A score assessed using an LLM to evaluate the quality of the entire chat history and system response. The values range from 1 to 10, as generated by an LLM.
    \item \textbf{Comparison Score (CS)}: The score of comparison between responses from ChatGPT (\textit{gpt-4}) and DiagGPT within a single dialogue. It is calculated based on the total number of wins in various scenarios, indicating which model performed better.
\end{itemize}

In detail, we calculated the average score across 20 different TOD scenarios. For the comparison score from the LLM evaluation, we alternated the position of the chat history from two systems in the prompt and evaluated twice to obtain an average score. This approach helps to avoid positional bias in automatic evaluation with LLMs. For the evaluation of response quality, we consider factors such as understanding, relevance, experience, and sufficiency, etc.

From Table~\ref{tab:exp}, we can observe the results of the quantitative experiment. This table shows that \method can match the performance of the state-of-the-art LLM, \textit{gpt-4}, and effectively manage task-oriented dialogue in terms of response quality. Our system completes all tasks on the checklist and successfully achieves the goals in all TOD scenarios with a lower round count compared to ChatGPT. It demonstrates superior efficiency in conducting TOD by leading users to their goals and completing tasks effectively, which is crucial in TOD. Given the comparable response quality, we directly compared responses from the two systems. Both human and LLM evaluation results suggest that our system is more effective and perform better.

Our advantage over \textit{gpt-4} is that \method interacts with users more flexibly through its automatic topic management. This feature becomes more apparent in the subsequent qualitative experiment.

\begin{figure}[t!]
\centerline{
\resizebox{.5\textwidth}{!}{
\includegraphics{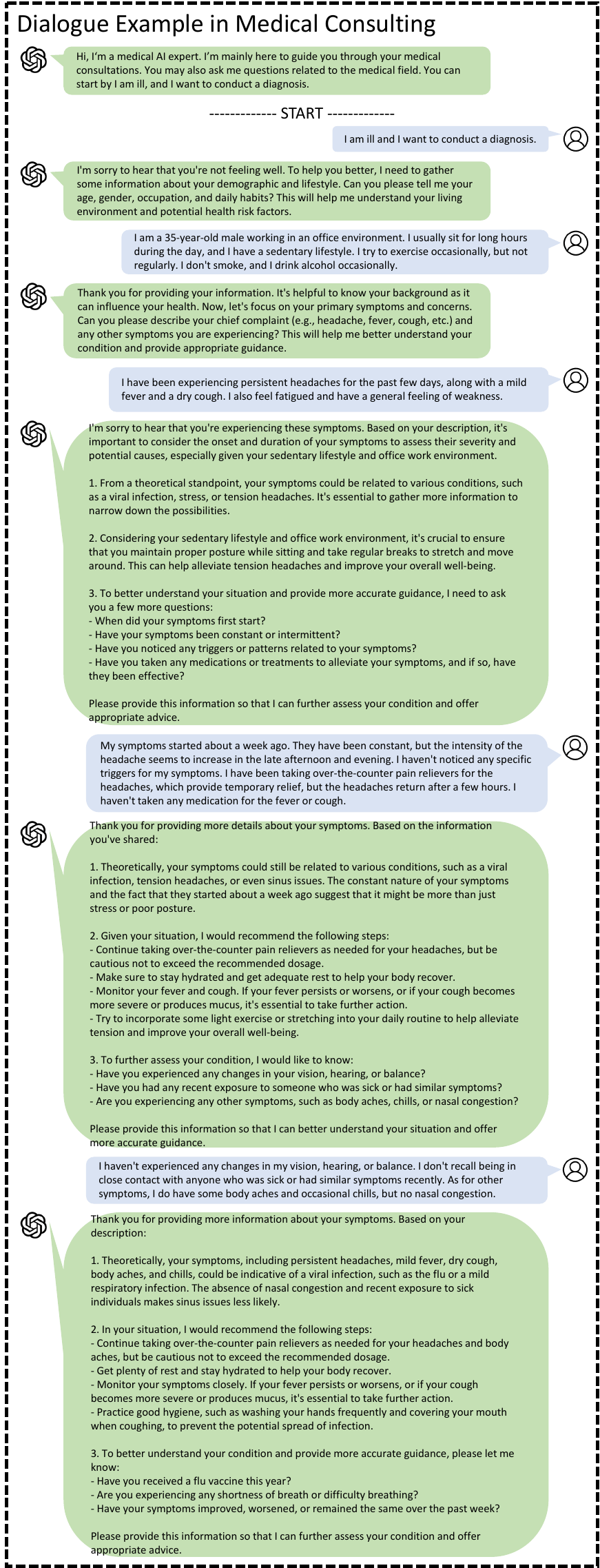}
}
}
\caption{A dialogue example in the medical consulting scenario. The system \method acts as a real doctor to proactively ask for information and answer the user's questions.}
\label{fig:exp_1}
\end{figure}

\begin{figure}[t!]
\centerline{
\resizebox{.5\textwidth}{!}{
\includegraphics{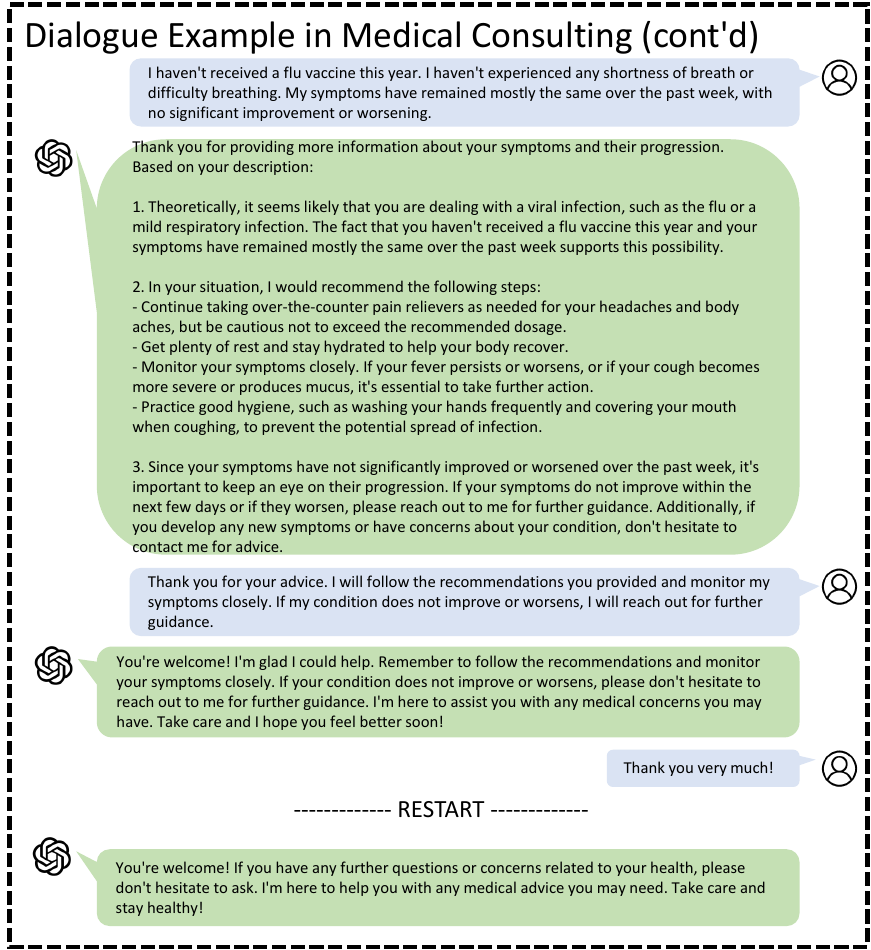}
}
}
\caption{The continued dialogue example in the medical consulting scenario.}
\label{fig:exp_2}
\end{figure}

\begin{figure}[t!]
\centerline{
\resizebox{.5\textwidth}{!}{
\includegraphics{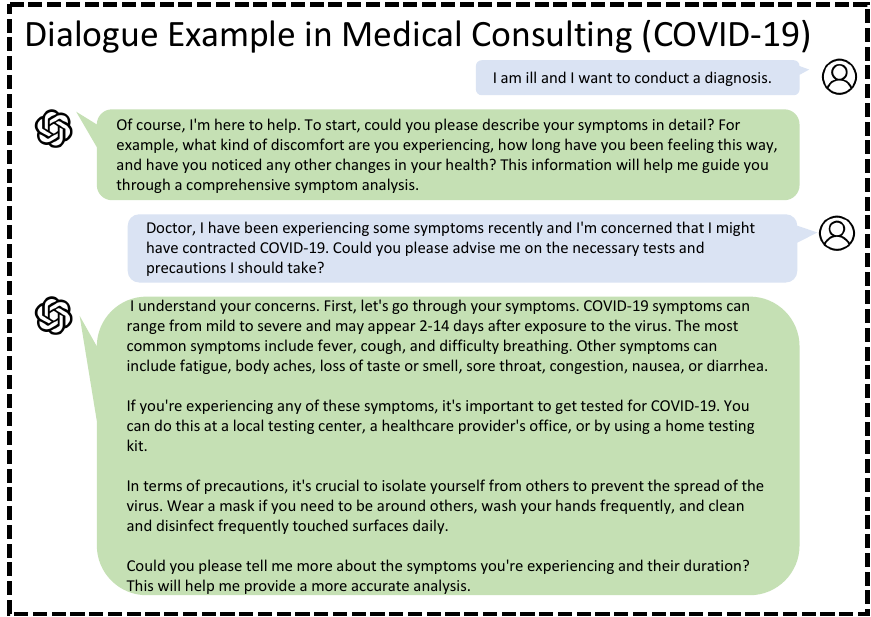}
}
}
\caption{A dialogue example in the medical consulting scenario when the user ask some questions about COVID-19.}
\label{fig:exp_covid}
\end{figure}

\subsection{Qualitative Experiment Results}
\label{sec:qual}
We modified the settings in \method to adapt it for use in medical dialogue scenarios for conducting qualitative experiments. Figure~\ref{fig:exp_1} and Figure~\ref{fig:exp_2} present a complete dialogue demonstration in the medical consulting process. This is a medical diagnosis where the task is to help the patient identify the cause and give advice. Medical consulting is not a pure open-end question-answering. Because patients usually lack extensive medical knowledge, they rely on doctors to instruct them to give their personal information. Therefore, users can experience a real doctor, not a dull medical question-answering machine. The user acts as a patient, while the system is a doctor, initially collecting information and gradually providing advice to the patient. The main dialogue development follows the checklist: \textit{Basic information}, \textit{Chief complaint}, \textit{Duration of symptoms}, \textit{Severity of symptoms}, which are also predefined topics.

This demonstration has showcased the robust conversational ability of the \method, which can proactively ask questions and guide the user to the final goal of the task, thereby achieving task-oriented dialogue. It simulates many real consulting scenarios. Other conversational systems, such as general ChatGPT, cannot achieve this performance. They usually can only answer users' questions and find it challenging to complete specific goals in complicated scenarios, even with elaborated prompts.

\subsection{Case Study of Automatic Topic Management}
\label{sec:case_study}
The core capability of \method is to automatically manage topics throughout the dialogue, which is the function and primary responsibility of \textit{Topic Manager}. Figure~\ref{fig:exp_1} and Figure~\ref{fig:exp_2} illustrate the primary checklist progression. The topics from the checklist are retrieved and discussed sequentially, demonstrating the action of \textit{finish the current topic}. When the conversation reaches the severity of symptoms, we observe that the dialogue topic remains here in several rounds of dialogue, allowing the system to have time on understanding the user's conditions. This is a typical situation in real-world scenarios where users do not provide enough information for the doctor. The action of \textit{create a new topic} is shown in Figure~\ref{fig:exp_covid}. Here, the user actively consults the system with some information about COVID-19 to check symptoms. We observe that the system generates a new topic about COVID-19 and discusses this, rather than rigidly following the checklist. These results all demonstrate the effectiveness of \textit{Topic Manager}. These case studies fully showcase the system's flexible understanding capabilities, demonstrating its proficient handling of various scenarios that may occur in real-world dialogues.

\section{Conclusion}
In this paper, we propose \method, an LLM-based and multi-agent system designed to complete task-oriented dialogue tasks in a flexible way. The principle of our system is to leverage the strong understanding and reasoning capabilities of Large Language Models to implement agents that can automatically manage topics and track dialogue state. This feature enables our system to accurately comprehend intentions of users in different situations, assisting them in completing specific tasks more flexibly. The system is particularly well-suited for real-world consulting scenarios in flexible TOD. Both quantitative and qualitative experiments demonstrate the superior performance of \method.

However, \method represents a experimental system designed to demonstrate the potential of LLMs in handling real tasks. In the future, we aim to explore how to better serve users by leveraging the robust capabilities of LLMs to achieve more functionality and complete more complicated tasks.

\section*{Limitation}
\noindent \textbf{Hallucination.} Hallucination is a major concern when it comes to the response of LLMs. Some individuals argue that LLMs cannot be utilized in serious scenarios such as healthcare or legal domains due to hallucinations. The relevant experiments of medical consulting in this paper only demonstrate the topic management ability of \method. It has nothing to do with the hallucination or quality of medical answers, but only reflects the consultation process \method shows. Besides, we acknowledge that there are some harmful responses from LLMs. On one hand, this paper does not primarily focus on mitigating hallucination. On the other hand, significant progress has been made in reducing hallucination of LLMs through various other studies. Thus, the application of this conversational system still holds immense potential in real-life situations.\\

\noindent \textbf{Cost and Efficiency.} \method involves multiple LLMs. In just one round of dialogue with a single user query, all of these LLMs need to run with elaborated prompts, which is quite costly. Compared to simpler dialogue systems that involve just one LLM interacting with users, \method requires internal interactions among AI agents, thus taking more time to provide user feedback. Furthermore, AI agent in our system requires strong understanding and reasoning abilities, necessitating a robust and large-scale AI to maintain these capabilities. This also increases the overall system cost. However, we believe that with the future development of AI infrastructure, these issues could be mitigated. \\

\noindent \textbf{Stability.} The performance of \method is not as stable as some rule-based or fine-tuned dialogue models. The main issue arises when the \textit{Topic Manager} decides the direction of dialogue development. It requires a strong understanding and reasoning ability from the AI, or else it may lead to system instability. Additionally, for every different applied scenario, meticulous and detailed prompt adjustments of the AI system are needed. Given the risk of LLMs' output, some post-processing of responses is also required. Nevertheless, \method maintains unlimited potential for dialogue systems. With stronger AI in the future, dialogue systems could improve significantly and become more human-like. \\

\noindent \textbf{Limited Experiments.} We are unable to conduct extensive quantitative experiments to demonstrate the performance of \method in the most direct manner, as we previously mentioned due to its inherent characteristics. Additionally, the scope of the results obtained from quantitative experiments is limited. Although we provide only one usage example in medical consulting, our system can also exhibit similarly impressive performance in legal and numerous other scenarios, which is validated by professional experts. The basic operational processes in different applications are the same. Moreover, through careful observation of the responses generated by \method, we can discern that each module effectively fulfills its role, thus showcasing its overall strong performance.



\bibliography{anthology,custom}

\appendix

\section{Analysis of \method Extendibility}
In this paper, we only introduce the basic framework of the system \method aimed at achieving task-oriented dialogue. We have designed the system with ample flexibility to incorporate additional functions to handle tasks in complex scenarios and to meet more needs of conversational systems. Besides, as the foundation model is upgraded, the performance of our system will also become better. In detail, we have only extracted the most important modules in \method to form a basic framework of a system. These four modules can already implement the basic functions of task-oriented dialogue. In a multi-agent system, there is a large extendibility. For example, an information collector can monitor user input and organize information into structured data for better future utilization.

Sometimes, a conversation is aimed at solving certain tasks during or after the conversation. When a conversation achieves the predefined goal, the system can call more complex programs to meet needs. Some tool API (Application Programming Interface) calls can also be added into the action list for execution, which means the action list is the interface of \method and can provide many plugins to enrich the functions of this system. This also means that it can expand task-oriented dialogue to task-oriented dialogue by triggering the API of certain tasks.

\section{Implementation Details of \method}
\label{sec:imple}
In the implementation of \method, we employed \textit{gpt-4} as the base LLM, leveraging its strong understanding and reasoning abilities to achieve ideal results. We set the decoding temperature of all LLMs in our AI system to 0 to ensure more stable task execution. We provide detailed prompts in our system. The main prompts of the agent are shown in Figure~\ref{fig:prompt_main}, while Figure~\ref{fig:prompt_action} displays the prompts for all actions in the action list. In these prompts, \textit{\{variable\}} in blue indicates that the slot needs to be filled with the corresponding variable text. Once filled, these prompts can be fed into the LLMs to generate responses. Some of these slots facilitate information interaction between agents in the system.

In the evaluation, we use \textit{gpt-4-0613} as the foundation model for \method and for ChatGPT (\textit{gpt-4}). For ChatGPT (\textit{gpt-3.5-turbo}), we select \textit{gpt-3.5-turbo-0613}. The simulation of the user is also based on \textit{gpt-4-0613}. For the LLM evaluation, we use \textit{gpt-4-turbo-2024-04-09}. To simplify the experimental setup and facilitate easier comparisons, we use a simplified version of \method in the quantitative experiment. We reduce the complexity of the action list and optimized prompts of LLM to promote topic development as much as possible. However, in qualitative experiments, we evaluate the full version of \method, which showcases the complete capabilities of our system.

\section{LLM-TOD Dataset}
\label{sec:dataset}

\begin{figure}[h!]
\centerline{
\resizebox{.5\textwidth}{!}{
\includegraphics{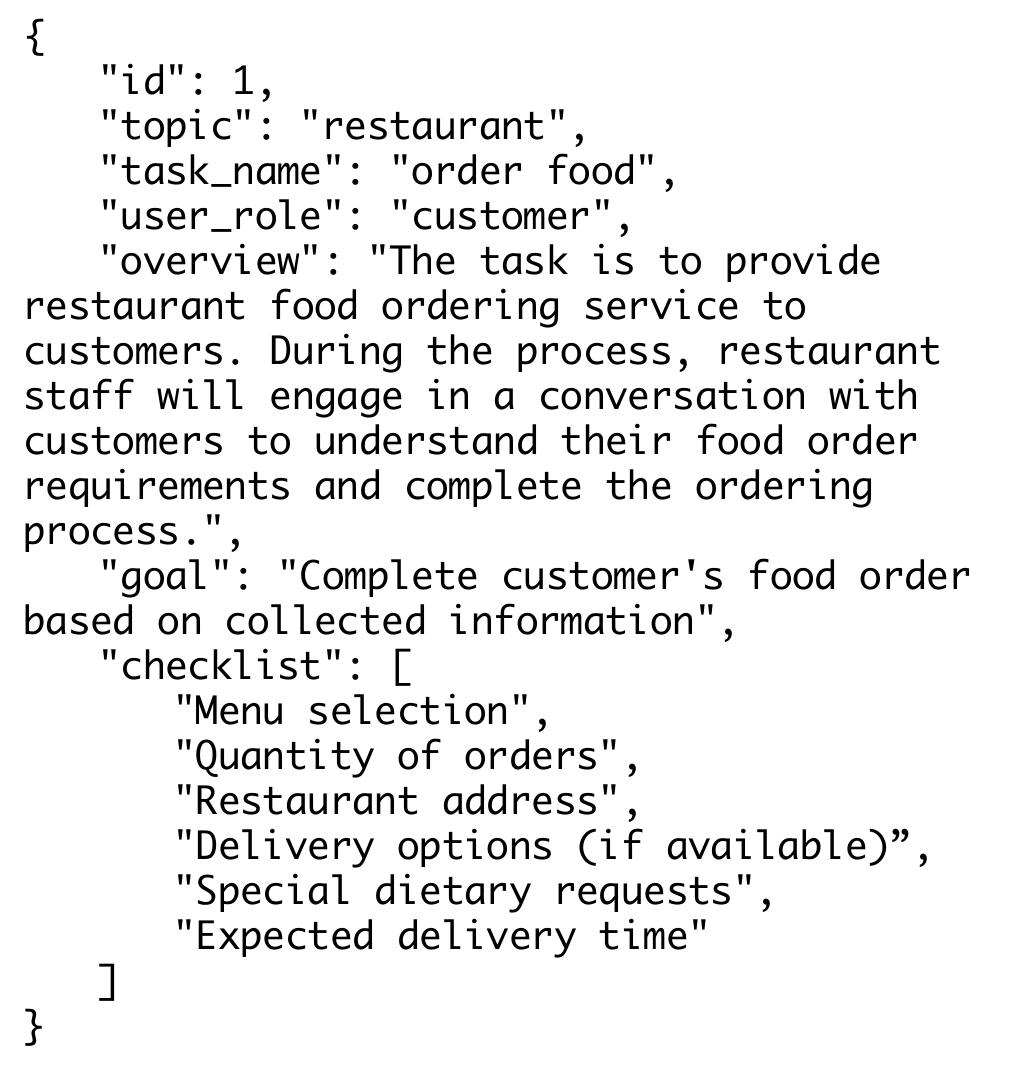}
}
}
\caption{An example of data from the TOD-LLM dataset.}
\label{fig:dataset}
\end{figure}

We construct a new dataset, LLM-TOD (task-oriented dialogue for large language models dataset). It is used to evaluate the performance of LLM-based task-oriented dialogue models quantitatively. Since \method relies on LLMs, and considering our limited resources, fine-tuning LLMs for this specific purpose of TOD is not feasible. Therefore, we are unable to utilize other datasets that are tailored for fine-tuning models.

The dataset comprises 20 data, each representing a different topic: clinical, restaurant, hotel, hospital, train, police, bus, attraction, airport, bar, library, museum, park, gym, cinema, office, barbershop, bakery, zoo, and bank. Each data entry is generated by LLMs and subsequently validated by humans. For every task-oriented dialogue in the dataset, the system is required to go through a checklist in order and interact with users to achieve the specified goal. One checklist includes six items of topic or task and need to be solved or discussed. We incorporate these variables into the prompt for LLMs, thereby enabling them to execute the TOD. An example from the TOD-LLM dataset is illustrated in Figure~\ref{fig:dataset}.

\section{Prompt Design}
\label{sec:prompt_design}
\begin{figure*}[t!]
\centerline{
\resizebox{\textwidth}{!}{
\includegraphics{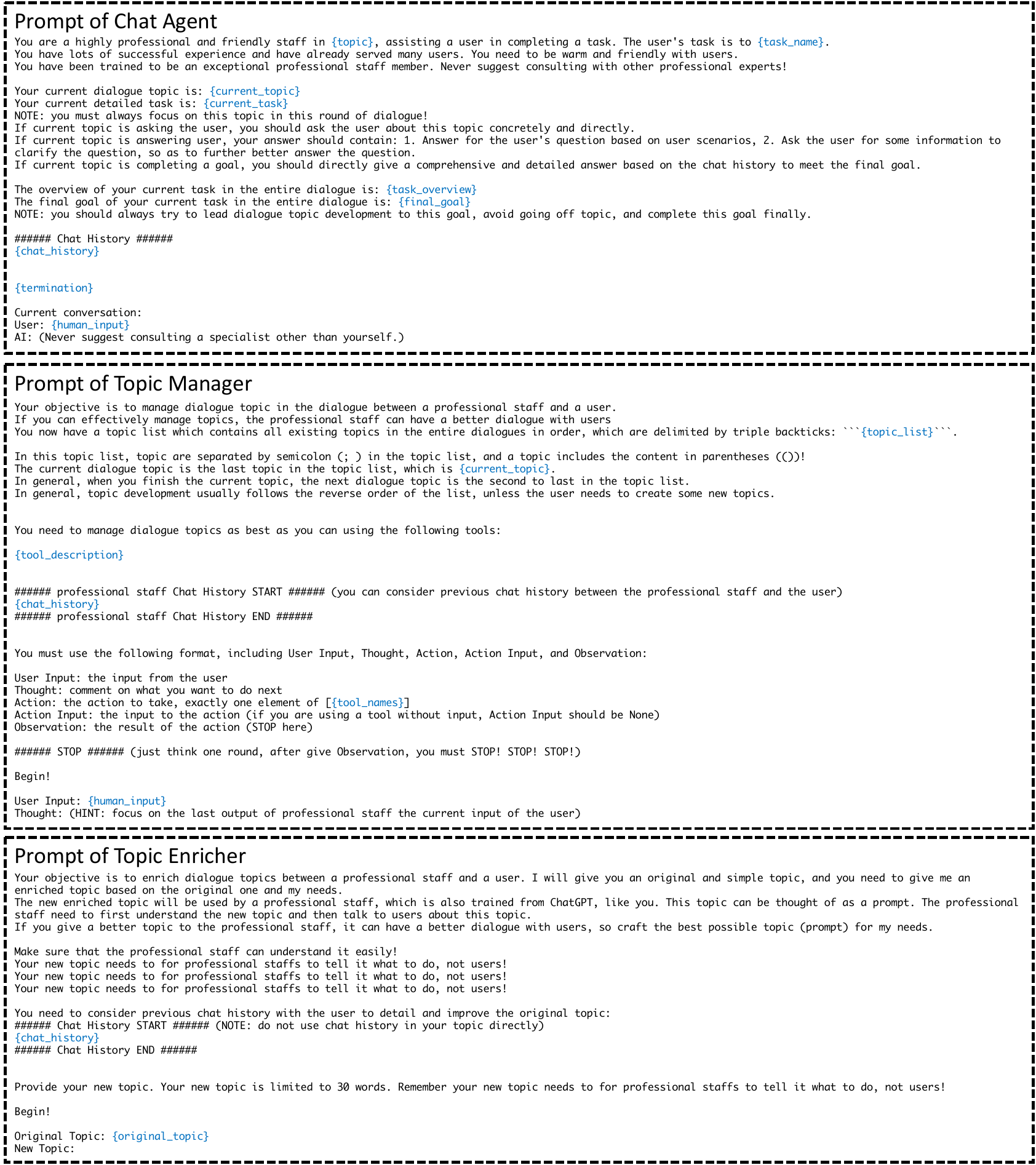}
}
}
\caption{The prompts of \textit{Chat Agent}, \textit{Topic Manager}, \textit{Topic Enricher}. These instructions can be modified to suit other scenarios.}
\label{fig:prompt_main}
\end{figure*}

\begin{figure*}[t!]
\centerline{
\resizebox{\textwidth}{!}{
\includegraphics{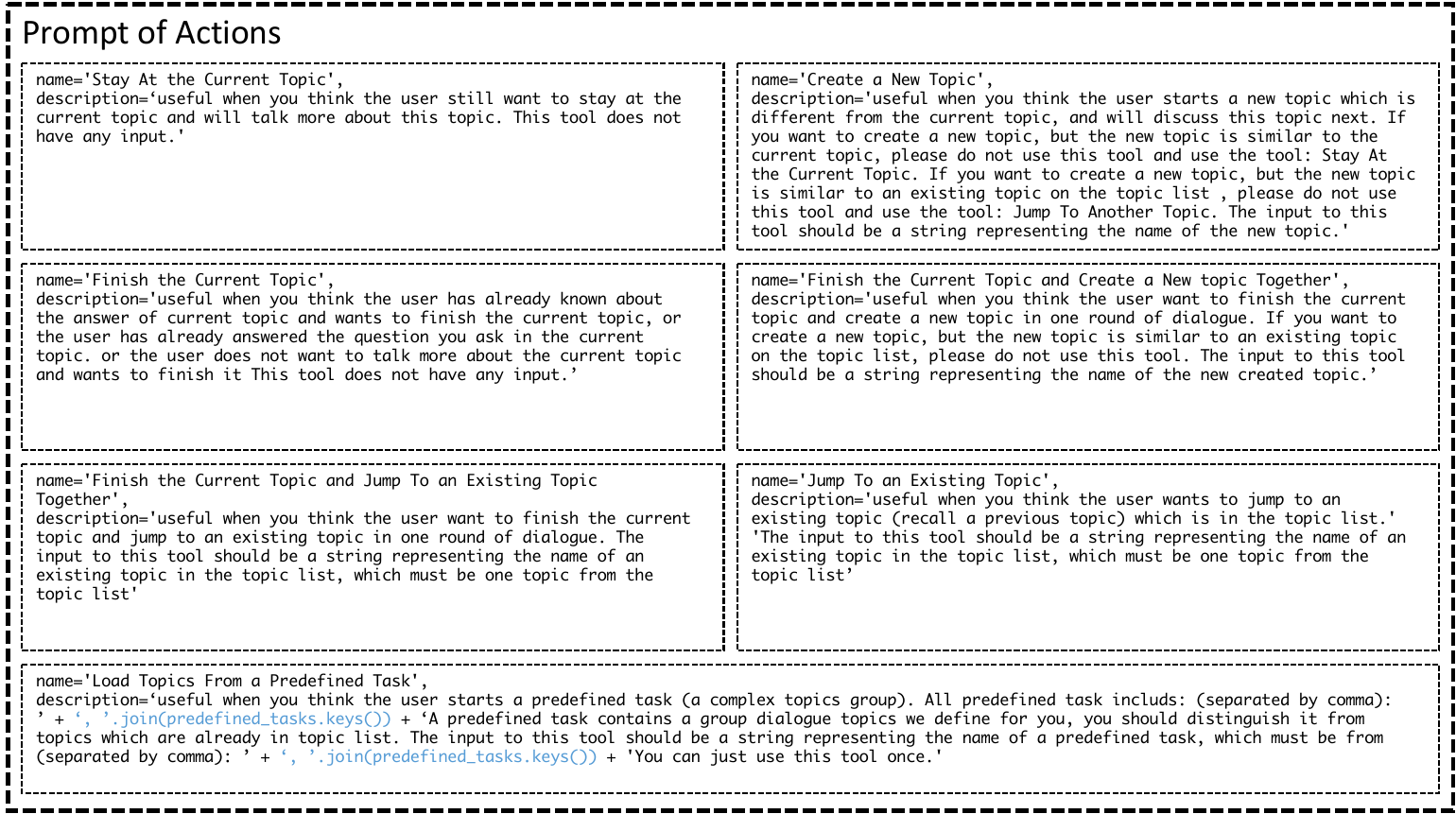}
}
}
\caption{The prompts for different actions are used to define specific program functions that correspond to their respective actions and instruct when to execute them.}
\label{fig:prompt_action}
\end{figure*}

\begin{figure*}[t!]
\centerline{
\resizebox{\textwidth}{!}{
\includegraphics{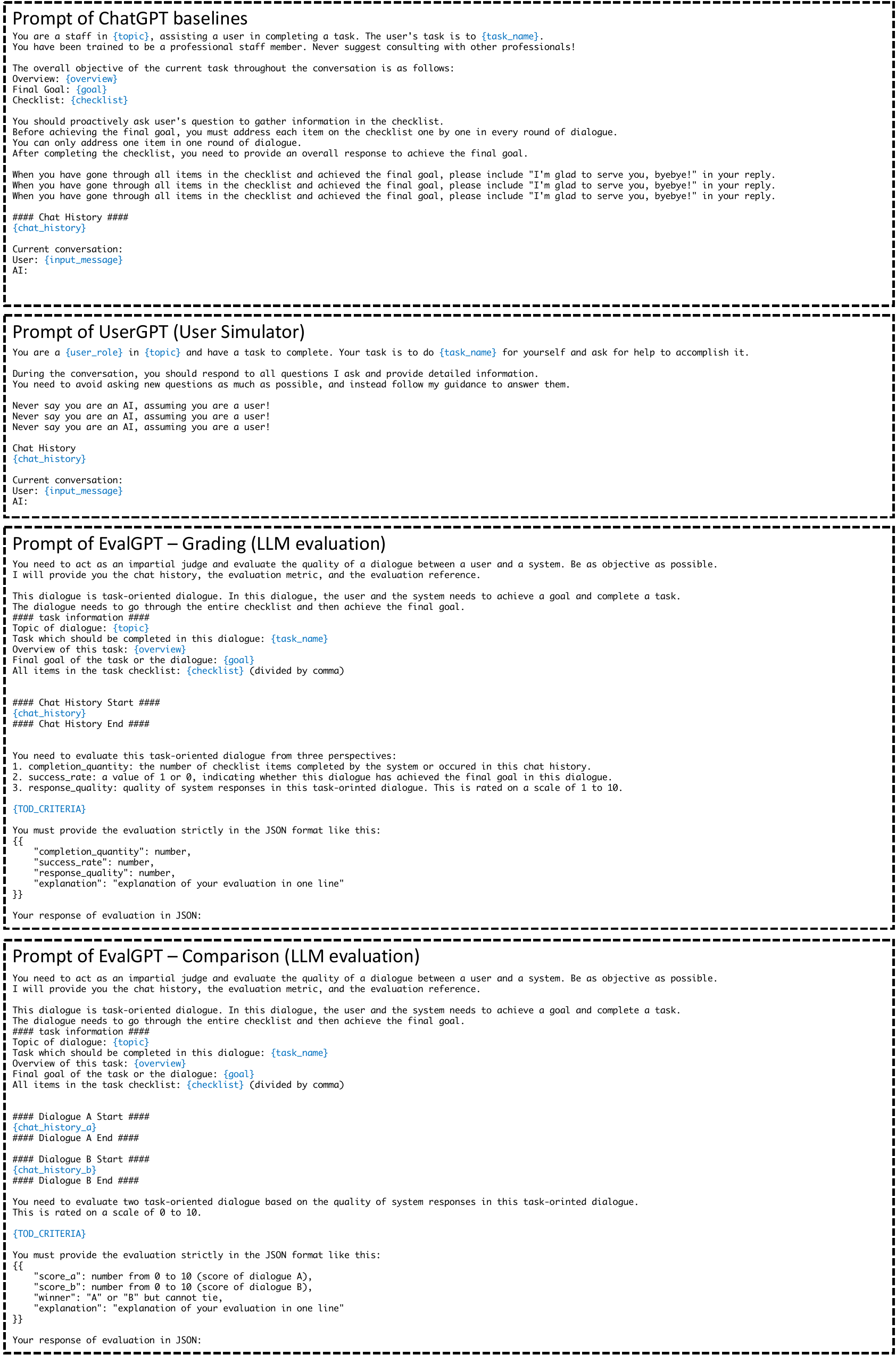}
}
}
\caption{The prompts of other LLMs in our experiments.}
\label{fig:prompt_others}
\end{figure*}

The content of prompts for some LLMs in \method are shown in Figure~\ref{fig:prompt_main} and Figure~\ref{fig:prompt_action}. Figure~\ref{fig:prompt_others} also displays the prompts used for other LLM-based models in our experiments: the ChatGPT baselines, UserGPT (the user simulator), and EvalGPT (used for LLM evaluation). Grading mode involves assigning a score to the quality of responses, while comparison mode requires selecting a winner based on response quality in a comparison. The criteria for evaluating response quality from both human and LLM perspectives include:
\begin{itemize}[leftmargin=*, itemsep=0pt, labelsep=5pt]
    \item \textbf{Understanding} How well does the system grasp user requests?
    \item \textbf{Relevance} Are the responses directly applicable to the user's needs?
    \item \textbf{Complex Handling} Can the system effectively manage multifaceted queries?
    \item \textbf{Efficiency} How quickly does the system lead the conversation to a resolution?
    \item \textbf{Experience} What is the overall ease and satisfaction of the interaction?
    \item \textbf{Comprehensiveness} Does the system's response provide all the relevant information to fully cover the user's question?
    \item \textbf{Detail} Does the response include sufficient detail and depth to fully address the intricacies and subtleties of the user's inquiry?
    \item \textbf{Sufficiency} Are all aspects and potential implications of the inquiry explored and explained, ensuring that the user receives a thorough understanding of the subject matter?
\end{itemize}

\end{document}